\pdfoutput=1

\documentclass[11pt]{article}

\usepackage[preprint]{acl}

\usepackage{times}
\usepackage{latexsym}

\usepackage[T1]{fontenc}

\usepackage[utf8]{inputenc}

\usepackage{microtype}

\usepackage{inconsolata}

\usepackage{graphicx}
\usepackage{hyperref}
\usepackage{url}
\usepackage{booktabs}
\usepackage{multirow}
\usepackage{amsmath}
\usepackage{enumitem}
\usepackage{subcaption}
\usepackage{float}

\definecolor{mplgreen}{HTML}{2CA02C}
\definecolor{mlppurple}{HTML}{800080}

\title{Predicting Task Performance with Context-aware Scaling Laws}

\author{Kyle Montgomery\textsuperscript{1*}, David Park\textsuperscript{2*}, Jianhong Tu\textsuperscript{1}, \\\textbf{Michael Bendersky\textsuperscript{3}, Beliz Gunel\textsuperscript{4}, Dawn Song\textsuperscript{5}, Chenguang Wang\textsuperscript{1}$^\dagger$}\\
\textsuperscript{1} UC Santa Cruz, \textsuperscript{2} Washington University in St. Louis, \textsuperscript{3}Databricks, \\ \textsuperscript{4}Google DeepMind, \textsuperscript{5}UC Berkeley\\
\texttt{\{kylemontgomery, chenguangwang\}@ucsc.edu}}

\begin{document}
\maketitle

\def\thefootnote{$^*$}\footnotetext{Equal contribution.}
\def\thefootnote{$^\dagger$}\footnotetext{Corresponding author.}

\begin{abstract}
Scaling laws have transformed our understanding of large language models by linking upstream metrics like cross-entropy loss to design factors such as model size, training data, and compute. However, these conventional laws fail to capture downstream task performance, where context plays a critical role. In this work, we propose a straightforward, interpretable framework that jointly models downstream performance as a function of the training compute and the provided context. We empirically validate our framework by fitting it on the observed downstream performance of extended-context variants of Llama-2-7B and Llama-2-13B across 65,500 unique instances spanning three tasks: arithmetic reasoning, common sense reasoning, and machine translation. Our results demonstrate that our framework accurately models in-distribution downstream performance, generalizes across three orders of magnitude in training compute, and reliably extrapolates performance as the amount of context increases. These findings offer valuable insights into the interplay between training compute and context utilization, providing guidance for designing more efficient long-context LLMs for diverse downstream tasks. Our code is available at \url{https://github.com/wang-research-lab/context-scaling}.
\end{abstract}

\section{Introduction}
Neural scaling laws~\citep{hestness2017deeplearningscalingpredictable, kaplan2020scalinglawsneurallanguage}, which describe how model performance scales with the number of model parameters, the size of the training dataset, or the amount of training compute, have shaped our understanding of how large language models (LLMs)~\citep{brown2020languagemodelsfewshotlearners, touvron2023llama2openfoundation, gemmateam2024gemma2improvingopen, grattafiori2024llama3herdmodels, openai2024gpt4technicalreport} improve with increased resources. These findings have guided the design and development of increasingly larger models, providing a blueprint to optimally scale up performance under a fixed compute budget~\citep{hoffmann2022trainingcomputeoptimallargelanguage, openai2024gpt4technicalreport}. 

While upstream metrics like cross-entropy loss serve as convenient proxies during model development, in real-world applications, downstream performance often diverges from these upstream trends~\citep{wei2022emergentabilitieslargelanguage, hu2024predictingemergentabilitiesinfinite}. Accurate upfront performance estimates for downstream tasks can help guide model development and identify emergence or saturation on certain tasks with fewer costly experiments. Existing works on predicting downstream performance often rely on overly complicated, less interpretable methods. For instance,~\citet{chen2024scalinglawspredictingdownstream} utilizes a two-stage approach using upstream loss as an intermediary, while~\citet{ye2023predictablelargelanguagemodel} fits a multi-layered perceptron to predict performance on BIG-Bench~\citep{srivastava2023imitationgamequantifyingextrapolating}.

\begin{figure}
    \centering
    \includegraphics[width=\linewidth]{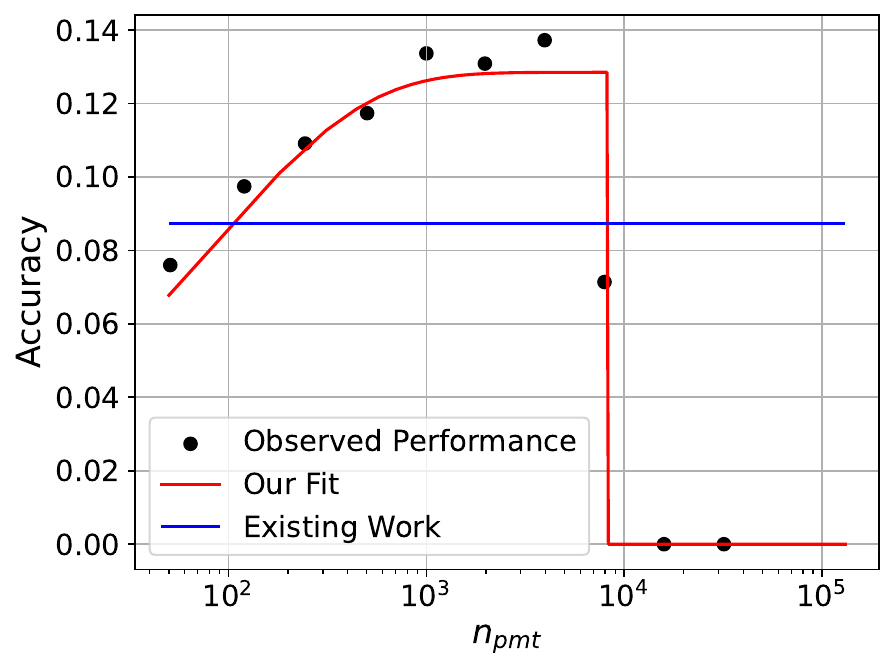}
    \caption{Existing approaches ignore the impact of context length and predict an average performance level regardless of the number of in-context demonstrations. In comparison, our context-aware fit closely tracks the observed performance as additional context is provided.}
    \label{fig:fit-comparison}
\end{figure}

In contrast, we propose a straightforward, interpretable framework that directly models the downstream performance of LLMs across a number of tasks. The key is to jointly model downstream performance as a function of the training compute and the provided context. Specifically, we develop a functional form (see Eq.~\eqref{eq:1}) which combines two saturating power-law terms (one in the amount of training compute and another in the amount of context) along with a penalty term to account for cases in which the context exceeds the model's context limit. This formulation is motivated by the intuition that downstream performance improves with increased training compute and longer, yet relevant, context until the benefits saturate or the context limit is exceeded. Figure~\ref{fig:fit-comparison} compares our fit to existing methods that do not consider context.

We empirically validate our scaling framework by fitting it on the observed downstream performance of extended-context variants of Llama-2-7B and Llama-2-13B~\citep{touvron2023llama2openfoundation, peng2023yarnefficientcontextwindow} across 65,500 unique instances spanning three tasks: arithmetic reasoning, common sense reasoning, and machine translation. Our results demonstrate that our framework accurately predicts downstream performance for both Llama‑2‑7B and Llama‑2‑13B (Sec.~\ref{sec:results-downstream}). Furthermore, we find that our fits generalize well on held-out models spanning 3 orders of magnitude in training compute (Sec.~\ref{sec:generalization-compute}). Similarly, we demonstrate that our fits generalize to longer contexts, even as the context exceeds a model's context limit (Sec.~\ref{sec:generalization-context}). Lastly, we show that our fits generalize across different context-extension techniques (Sec.~\ref{sec:extension-technique}). These findings offer valuable insights into the interplay between training compute and context utilization, providing guidance for designing more efficient long-context LLMs for diverse downstream tasks.\\

\noindent Our main contributions are threefold:
\begin{itemize}[itemsep=0.25pt]
    \item We propose a framework that extends conventional neural scaling laws to downstream tasks by incorporating the context length and context limit, providing a more accurate model of LLM performance across varying context lengths.
    \item We empirically fit this framework to Llama-2 models with extended context windows across 3 tasks: arithmetic reasoning, common sense reasoning, and machine translation. We demonstrate the generality of our approach by showing that our scaling laws hold across 3 orders of magnitude in training compute, 4 orders of magnitude in context length, and across different context-extension techniques. 
    \item Our framework offers an interpretable tool for understanding the interplay between compute, context, and downstream performance, providing insights that can guide the design of future long-context LLMs.
\end{itemize}

\section{Background}
Here, we introduce relevant preliminaries, including notation conventions and the process of extending the context window of the Llama-2 models~\citep{touvron2023llama2openfoundation}. 

\subsection{Notation}
\label{sec:notation}
We adopt the following notation:
\begin{itemize}
    \item $\mathcal{P}$ -- aggregate performance on a downstream task. Occasionally, we'll use a subscript to denote the specific task (e.g., $\mathcal{P}_{\text{MT}}$ for machine translation). 
    \item $N$ -- the number of model parameters, excluding embedding/de-embedding parameters.
    \item $D$ -- the number of tokens in the training dataset.
    \item $C$ -- the amount of non-embedding training compute. Following~\citet{kaplan2020scalinglawsneurallanguage}, we estimate $C \approx 6N$ FLOPs per training token, or $C \approx 6ND$ FLOPs in total.
    \item $n_{\text{ctx}}$ -- the context limit of a model in tokens, i.e., the maximum number of positional embeddings computed for any training sequence. Often, we quote numerical values using $k$ to denote units of 1024 tokens. For example, a context limit of ``128k'' corresponds to $128 \times 1024 = 131072$ tokens.
    \item $n_{\text{pmt}}$ -- the length (in tokens) of a given input query or context. For simplicity, $n_{\text{pmt}}$ does not include generated/outputted tokens. 
\end{itemize}

\subsection{Extending Llama-2's Context Limit}
\label{sec:context-extension}

\begin{table*}[t]
    \centering
    \small
    \begin{tabular}{lcccc}
        \toprule
        Base Model & Non-embedding Params $\left(N\right)$ & Context Limit $\left(n_{\text{ctx}}\right)$ & Dataset Size $\left(D\right)$ & Training Compute $\left(C\right)$\\
        \midrule
        \multirow{6}{*}{Llama-2-7B}   & \multirow{6}{*}{6,476,271,616}  & 4k   & 2.0T            & $7.7719 \times 10^{22}$ \\
                                      &                                 & 8k   & 2.0T $+$ 0.210B & $7.7723 \times 10^{22}$ \\
                                      &                                 & 16k  & 2.0T $+$ 0.419B & $7.7732 \times 10^{22}$ \\
                                      &                                 & 32k  & 2.0T $+$ 0.836B & $7.7748 \times 10^{22}$ \\
                                      &                                 & 64k  & 2.0T $+$ 1.678B & $7.7780 \times 10^{22}$ \\
                                      &                                 & 128k & 2.0T $+$ 3.355B & $7.7846 \times 10^{22}$ \\
        \midrule
        \multirow{6}{*}{Llama-2-13B}  & \multirow{6}{*}{12,688,184,320} & 4k   & 2.0T            & $1.5227 \times 10^{23}$ \\
                                      &                                 & 8k   & 2.0T $+$ 0.210B & $1.5227 \times 10^{23}$ \\
                                      &                                 & 16k  & 2.0T $+$ 0.419B & $1.5229 \times 10^{23}$ \\
                                      &                                 & 32k  & 2.0T $+$ 0.836B & $1.5232 \times 10^{23}$ \\
                                      &                                 & 64k  & 2.0T $+$ 1.678B & $1.5239 \times 10^{23}$ \\
                                      &                                 & 128k & 2.0T $+$ 3.355B & $1.5251 \times 10^{23}$ \\
        \bottomrule
    \end{tabular}
    \caption{The 12 checkpoints against which we fit scaling curves. The 4k variants are the official Llama-2-7B and Llama-2-13B checkpoints. The additional training tokens and compute from extending the context limit via YaRN~\citep{peng2023yarnefficientcontextwindow} are factored into $D$ and $C$.}
    \label{tab:models}
\end{table*}

Because the complexity of the self-attention layers grows quadratically in the sequence length~\citep{keles2022computationalcomplexityselfattention, dao2022flashattentionfastmemoryefficientexact}, LLMs are commonly pre-trained on short sequences (e.g., 4k tokens) rather than long sequences (e.g., 128k tokens). As a result, LLMs struggle to generalize to sequences longer than those seen during pre-training. Because we plan to explore how downstream performance varies with context length, Llama-2's original context limit of 4k tokens will not be sufficient. Fortunately, a number of techniques have been proposed that can extend the context window of LLMs for a fraction of the pre-training compute budget~\citep{chen2023extendingcontextwindowlarge, peng2023yarnefficientcontextwindow, xiong2023effectivelongcontextscalingfoundation}.

YaRN~\citep{peng2023yarnefficientcontextwindow} is our method of choice for extending Llama 2's context limit. We selected YaRN due to its high compute efficiency and strong empirical results compared to other techniques. YaRN involves fine-tuning the pre-trained model for a limited number of steps on sequences exceeding the pre-trained LLM's context limit in order to increase the effective size of the LLM's context limit so that it may better model long sequences. 

We adopt the methodology from~\citet{peng2023yarnefficientcontextwindow} and fine-tune Llama-2-7B and Llama-2-13B~\citep{touvron2023llama2openfoundation} for 400 steps with a global batch size of 64 on sequences of length $n'_{\text{ctx}}$ (where $n'_{\text{ctx}} > n_{\text{ctx}}$) from the PG-19 corpus~\citep{rae2019compressivetransformerslongrangesequence}. We use the AdamW optimizer~\citep{loshchilov2019decoupledweightdecayregularization} with $\beta_1=0.9$ and $\beta_2=0.95$, and a learning rate of $2 \times 10^{-5}$. We train variants of Llama-2-7B and Llama-2-13B with $n_{\text{ctx}} \in \{8\text{k}, 16\text{k}, 32\text{k}\}$, and source checkpoints for $n_{\text{ctx}} \in \{64\text{k}, 128\text{k}\}$ from~\citet{peng2023yarnefficientcontextwindow}.

In order to validate the effectiveness of the context extension training, we evaluate the performance of our 12 Llama-2 models in Table~\ref{tab:models} on RULER~\citep{hsieh2024rulerwhatsrealcontext}, a synthetic needle-in-a-haystack benchmark developed to evaluate long-context LLMs. Specifically, we evaluate each model on 100 instances per length, for each of RULER's 13 tasks. Results are displayed in Table~\ref{tab:ruler} and suggest that context extension via YaRN~\citep{peng2023yarnefficientcontextwindow} is somewhat effective. Interestingly, models tend to underperform when evaluated at their extended context limit, suggesting that training with a context limit well beyond the target evaluation range can lead to improved performance within that desired range.

\begin{table*}
    \centering
    \small
    \begin{tabular}{lccccccc}
    \toprule
    Model & $n_\text{ctx}$ & $n_\text{pmt} = 4\text{k}$ & $n_\text{pmt} = 8\text{k}$ & $n_\text{pmt} = 16\text{k}$ & $n_\text{pmt} = 32\text{k}$ & $n_\text{pmt} = 64\text{k}$ & $n_\text{pmt} = 128\text{k}$\\ 
    \midrule
    \multirow{6}{*}{Llama-2-7B} & 4k  & {0.822} & {0.000} & {0.000} & {0.000} & {0.000} & {0.000} \\
                                & 8k  & {0.829} & {0.586} & {0.000} & {0.000} & {0.001} & {0.005} \\
                                & 16k & {0.795} & {0.58} & {0.378} & {0.000} & {0.000} & {0.002} \\
                                & 32k & {0.746} & {0.599} & {0.517} & {0.317} & {0.000} & {0.000} \\
                                & 64k & {0.794} & {0.647} & {0.593} & {0.530} & {0.225} & {0.000} \\
                                & 128k & {0.776} & {0.663} & {0.552} & {0.439} & {0.383} & {0.129} \\ 
    \midrule
    \multirow{6}{*}{Llama-2-13B} & 4k & {0.861} & {0.000} & {0.000} & {0.000} & {0.000} & {0.000} \\
                                & 8k & {0.870} & {0.625} & {0.000} & {0.000} & {0.000} & {0.000} \\
                                & 16k & {0.865} & {0.679} & {0.392} & {0.000} & {0.000} & {0.000} \\
                                & 32k & {0.848} & {0.727} & {0.622} & {0.378} & {0.000} & {0.000} \\
                                & 64k & {0.860} & {0.734} & {0.612} & {0.511} & {0.282} & {0.001} \\
                                & 128k & {0.819} & {0.684} & {0.586} & {0.484} & {0.447} & {0.163} \\ 
    \bottomrule
    \end{tabular}
    \caption{Accuracy of our extended Llama-2 models on RULER~\citep{hsieh2024rulerwhatsrealcontext}.}
    \label{tab:ruler}
\end{table*}

\section{Method}
\label{context-aware-scaling-laws}
We posit that aggregate task performance $\mathcal{P}$ can be modeled as the product of two saturating power laws in $C$ and $n_{\text{pmt}}$, with a sigmoid penalty term for when $n_{\text{pmt}} > n_{\text{ctx}}$. This form provides a good fit for a range of tasks, including arithmetic reasoning, common sense reasoning, and machine translation tasks. Formally, we model $\mathcal{P}$ as

{\scriptsize
\begin{align}
\label{eq:1}
\mathcal{P}(C, n_\text{pmt}, n_\text{ctx}) &= 
\overbrace{\Bigl[ 1 - \exp\Bigl(- A\, \Bigl(\frac{C}{C^{c}}\Bigr)^{\alpha} \Bigr) \Bigr]}^{\text{Saturating term in } C} \\
&\times 
\underbrace{\Bigl[ 1 - \exp\Bigl(- B\, \Bigl(\frac{n_\text{pmt}}{n_\text{pmt}^c}\Bigr)^{\beta} \Bigr) \Bigr]}_{\text{Saturating term in } n_\text{pmt}} \times \underbrace{\sigma\left(n_\text{pmt} - n_\text{ctx}\right)}_{\text{Penalty term}} , \notag
\end{align}
}

\noindent where $A$, $C^c$, $\alpha$, $B$, $n_\text{pmt}^c$, and $\beta$ are parameters to be optimized. 

We select this form because we expect that the downstream performance $\mathcal{P}$ is proportional to diminishing terms in the amount of training compute $C$ (which integrates both model size $N$ and dataset size $D$)~\citep{chen2024scalinglawspredictingdownstream, owen2024predictablelanguagemodelbenchmark} and the context length~\citep{brown2020languagemodelsfewshotlearners, caballero2023brokenneuralscalinglaws}, assuming the context remains relevant as its length increases and $n_{\text{pmt}} \leq n_{\text{ctx}}$. We saturate these terms via exponentiation to ensure our predicted performance remains below the maximum theoretical performance of 1.0. The product form arises because compute and context are complementary, not additive; a significant lack in one dimension limits the benefit derived from the other. For example, providing more context is only beneficial to the extent that the model is capable of leveraging that additional context. We impose a sharp sigmoid penalty term because $\mathcal{P}$ is measured only on the generated tokens, and if $n_{\text{pmt}} > n_{\text{ctx}}$, then any generated tokens will fall beyond the range in which the model can make reliable predictions, meaning $\mathcal{P}$ degrades rapidly, especially on tasks that require extended and coherent generations (e.g., reasoning through a math word problem or translating an entire sentence).

\subsection{Datasets}
\label{sec:datasets}
We evaluate our 12 models in Table~\ref{tab:models} on 65,500 instances of varying lengths that span 3 tasks:
\begin{itemize}
    \item \textbf{Arithmetic reasoning} We collect 3550 testing instances across GSM8K~\citep{cobbe2021trainingverifierssolvemath}, MATH~\citep{hendrycks2021measuringmathematicalproblemsolving}, AQUA-RAT~\citep{ling2017programinductionrationalegeneration}, and Deepmind Math~\citep{saxton2019analysingmathematicalreasoningabilities}. Because the instances are rather short, we pack the context with up to 511 demonstrations sampled from the training splits of each dataset.
    \item \textbf{Common sense reasoning} We sample 1750 testing instances across PIQA~\citep{bisk2019piqareasoningphysicalcommonsense}, SIQA~\citep{sap2019socialiqacommonsensereasoningsocial}, OpenBookQA~\citep{mihaylov2018suitarmorconductelectricity}, HellaSwag~\citep{zellers2019hellaswagmachinereallyfinish}, WinoGrande~\citep{sakaguchi2019winograndeadversarialwinogradschema}, ARC-Easy/Challenge~\citep{clark2018thinksolvedquestionanswering}, and CommonSenseQA~\citep{talmor2019commonsenseqaquestionansweringchallenge}, and pack the context with up to 511 demonstrations from their respective training splits.
    \item \textbf{Machine translation} We sample 250 translation instances from WMT-14~\citep{bojar-EtAl:2014:W14-33} from each of German, French, Hindi, Czech, and Russian to English. As before, we pack the context with up to 511 demonstrations (of the same source language) and measure the BLEU-4~\citep{10.3115/1073083.1073135} score of the generation against the reference translation. 
\end{itemize}

Additional details can be found in Appendix~\ref{sec:dataset_details}.

\subsection{Fitting Procedure}
\label{sec:fitting-procedure}
For each task, we aggregate the results for each model by the context length, using the number of in-context demonstrations as a proxy for length. Within each group, we average over the context length and metric value for each instance. In doing so, we collect a number of records of the form ($C$, $n_\text{pmt}$, $n_\text{ctx}$, avg. metric value) on which we fit Eq.~\eqref{eq:1} for each of our 3 tasks.

\begin{table}[b]
    \small
    \centering
    \begin{tabular}{ccc}
    \toprule
    Parameter & Lower Bound & Upper Bound \\ \midrule
    $A$              &  0    &  100           \\
    $C^c$            &  0    &  $10^{30}$     \\
    $\alpha$         &  0    &  10            \\
    $B$              &  0    &  100           \\
    $n_\text{pmt}^c$ &  0    &  131,072        \\
    $\beta$          &  0    &  10            \\
    \bottomrule
    \end{tabular}
    \caption{Upper and lower bounds on $A$, $C^c$, $\alpha$, $B$, $n_\text{pmt}^c$, and $\beta$.}
    \label{tab:bounds}
\end{table}

\begin{figure*}[htbp]
  \centering
  \begin{subfigure}[b]{0.31\textwidth}
    \centering
    \includegraphics[width=\textwidth]{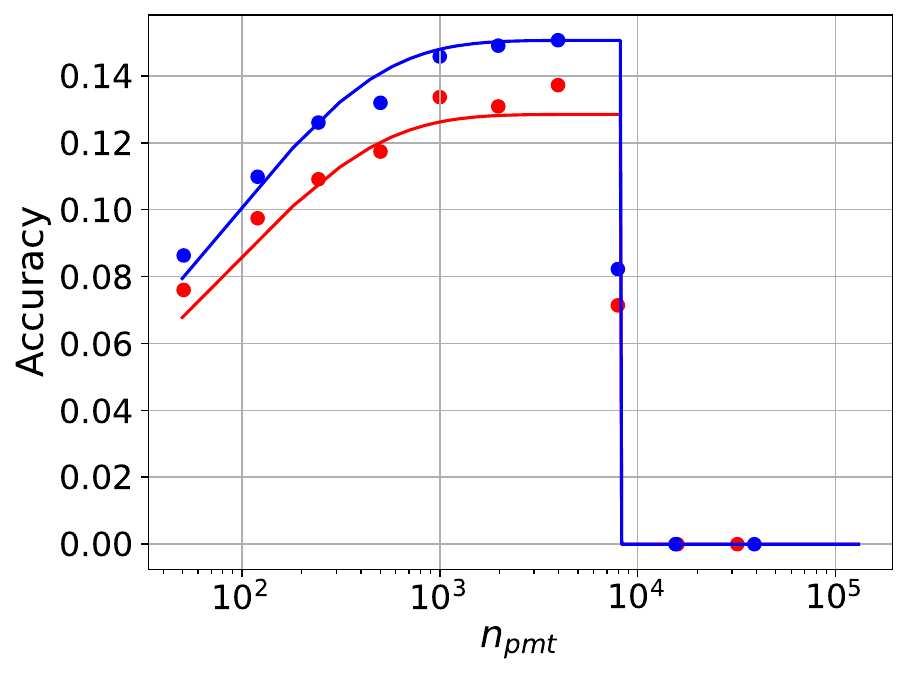}
    \caption{Arithmetic reasoning.}
    \label{fig:arithmetic_reasoning_2d}
  \end{subfigure}
  \hfill
  \begin{subfigure}[b]{0.31\textwidth}
    \centering
    \includegraphics[width=\textwidth]{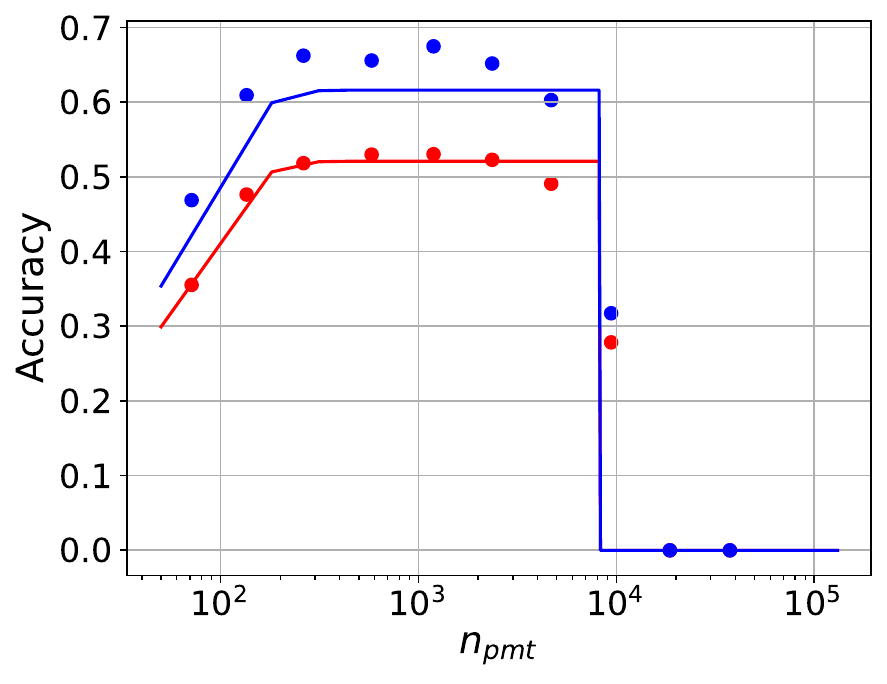}
    \caption{Common sense reasoning.}
    \label{fig:commonsense_reasoning_2d}
  \end{subfigure}
  \hfill
  \begin{subfigure}[b]{0.31\textwidth}
    \centering
    \includegraphics[width=\textwidth]{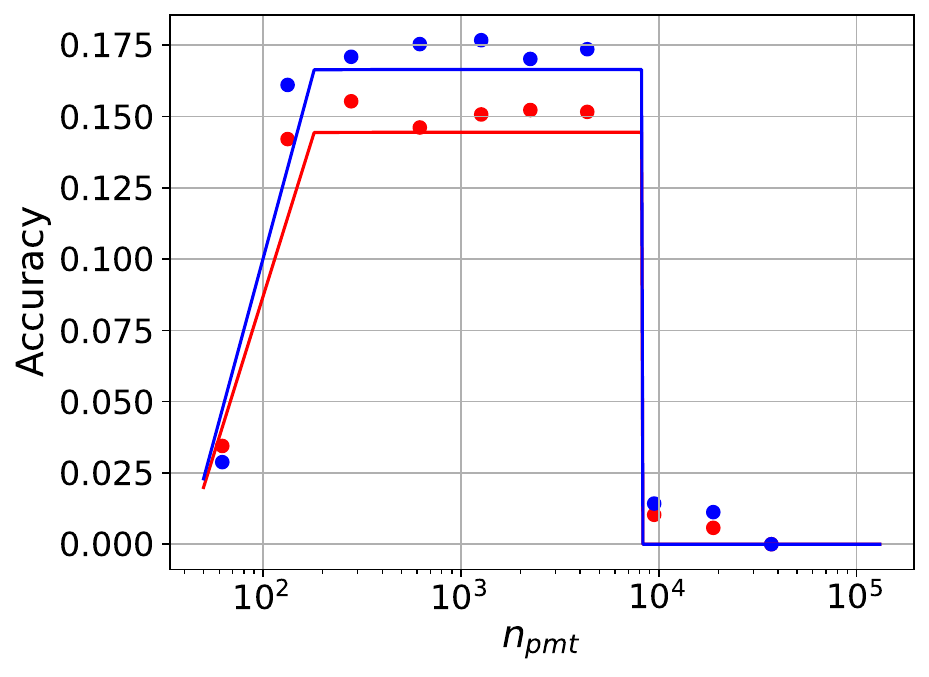}
    \caption{Machine translation.}
    \label{fig:machine_translation_2d}
  \end{subfigure}
  \caption{Contours of fits at $C=7.8 \times 10^{22}$ (\textcolor{red}{red}) and $C=1.5 \times 10^{23}$ (\textcolor{blue}{blue}) for $n_\text{ctx}=8\text{k}$ on three tasks: arithmetic reasoning (left), common sense reasoning (middle) and machine translation (right).}
  \label{fig:downstream}
\end{figure*}

\begin{table*}[htbp]
    \small
    \centering
    \begin{tabular}{lcccccc}
    \toprule
    Task & $A$ & $C^c$ & $\alpha$ & $B$ & $n_\text{pmt}^c$ & $\beta$\\
    \midrule
    Arithmetic reasoning   & $9.96$  & $9.7 \times 10^{29}$ & $0.26$ & $62.24$ & $1.3 \times 10^{5}$ & $0.56$ \\
    Common sense reasoning & $99.39$ & $1.5 \times 10^{28}$ & $0.40$ & $96.31$ & $3.5 \times 10^{3}$ & $1.12$ \\
    Machine translation    & $5.55$  & $5.4 \times 10^{29}$ & $0.23$ & $31.82$ & $3.0 \times 10^{2}$ & $2.97$ \\
    \bottomrule
    \end{tabular}
    \caption{Fits for $\mathcal{P}(C, n_{\text{pmt}}, n_{\text{ctx}})$ on 3 downstream tasks: arithmetic reasoning, common sense reasoning, and machine translation.}
    \label{tab:perf_fit}
\end{table*}

To fit the scaling curves, we use a two-stage optimization procedure that combines global search with local refinement. First, we use an out-of-the-box global optimizer to perform a broad search over the parameter space. Specifically, we use SciPy's \verb|differential_evolution| global optimization method, an evolutionary algorithm well suited for non-convex, non-linear optimization problems such as this~\citep{storn1997differential}. We define finite upper and lower bounds for each parameter, informed by~\citet{kaplan2020scalinglawsneurallanguage} and~\citet{xiong2023effectivelongcontextscalingfoundation}. We use the same bounds across all tasks, which are listed in Table~\ref{tab:bounds}. Finally, we do a pass through a local optimizer (SciPy's \verb|curve_fit|), using the estimate from the global optimizer as a starting point, to achieve a precise fit.

\section{Empirical Results}
\label{sec:results-downstream}

We model the aggregate performance $\mathcal{P}$ on each of our 3 tasks (arithmetic reasoning, common sense reasoning, and machine translation) using Eq.~\eqref{eq:1}. Unless otherwise noted, scaling laws are fit on the results of all 12 Llama-2 models in Table~\ref{tab:models} using the procedure outlined in Section~\ref{sec:fitting-procedure}. Table~\ref{tab:perf_fit} includes the parameter values that we found to be optimal for each task. Contours of our fits at $C=7.8 \times 10^{22}$ and $C=1.5 \times 10^{23}$ for $n_\text{ctx} = 8\text{k}$ are provided in Figure~\ref{fig:downstream}. Additional contours are provided in Appendix~\ref{sec:full-results}. We report the mean absolute prediction error $|\mathcal{P} - \hat{\mathcal{P}}|$, which is the average of residuals (in absolute value). When discussing individual residuals, we'll often include the sign of the residual to indicate the direction (i.e., whether we're under- or over-predicting).

On the arithmetic reasoning task, we achieve an excellent fit, with an average prediction error $|\mathcal{P}-\hat{\mathcal{P}}|$ of just 0.010. Similarly, on common sense reasoning and machine translation, we observe average prediction errors of 0.037 and 0.007, respectively. Additionally, we model the behavior around the boundary condition at $n_\text{pmt} = n_\text{ctx}$ surprisingly well.

\begin{table*}[htbp]
    \small
    \centering
    \begin{tabular}{lcccccc}
    \toprule
    Model & $C$ & $n_\text{ctx}$ & $\mathcal{P}_{\text{AR}} - \hat{\mathcal{P}}_{\text{AR}}$ & $\mathcal{P}_{\text{CSR}} - \hat{\mathcal{P}}_{\text{CSR}}$ & $\mathcal{P}_{\text{MT}} - \hat{\mathcal{P}}_{\text{MT}}$ \\
    \midrule
    Qwen-2.5-0.5B & $3.8 \times 10^{22}$ & 32k & +0.057 & +0.008 & -0.057  \\
    Gemma-2-2B    & $2.4 \times 10^{22}$ & 4k  & +0.066 & +0.260 & +0.059  \\
    Gemma-2-9B    & $4.0 \times 10^{23}$ & 4k  & +0.069 & +0.051 & +0.017  \\
    Gemma-2-27B   & $2.0 \times 10^{24}$ & 4k  & +0.024 & -0.099 & -0.054  \\
    Llama-2-70B   & $8.2 \times 10^{23}$ & 4k  & -0.002 & -0.031 & -0.025  \\
    \bottomrule
    \end{tabular}
    \caption{Generalization of fit on test models for arithmetic reasoning (AR), common sense reasoning (CSR), and machine translation (MT).}
    \label{tab:results_alt}
\end{table*}

\begin{figure*}[htbp]
  \centering
  \begin{subfigure}[b]{0.31\textwidth}
    \centering
    \includegraphics[width=\textwidth]{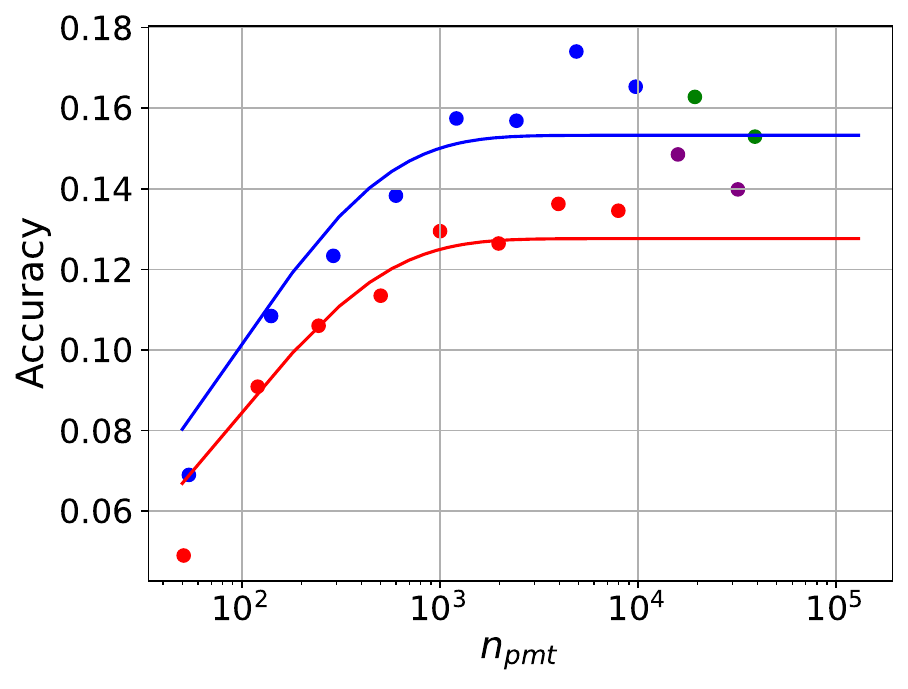}
    \caption{Arithmetic reasoning.}
    \label{fig:arithmetic_reasoning_2d_generalization}
  \end{subfigure}
  \hfill
  \begin{subfigure}[b]{0.31\textwidth}
    \centering
    \includegraphics[width=\textwidth]{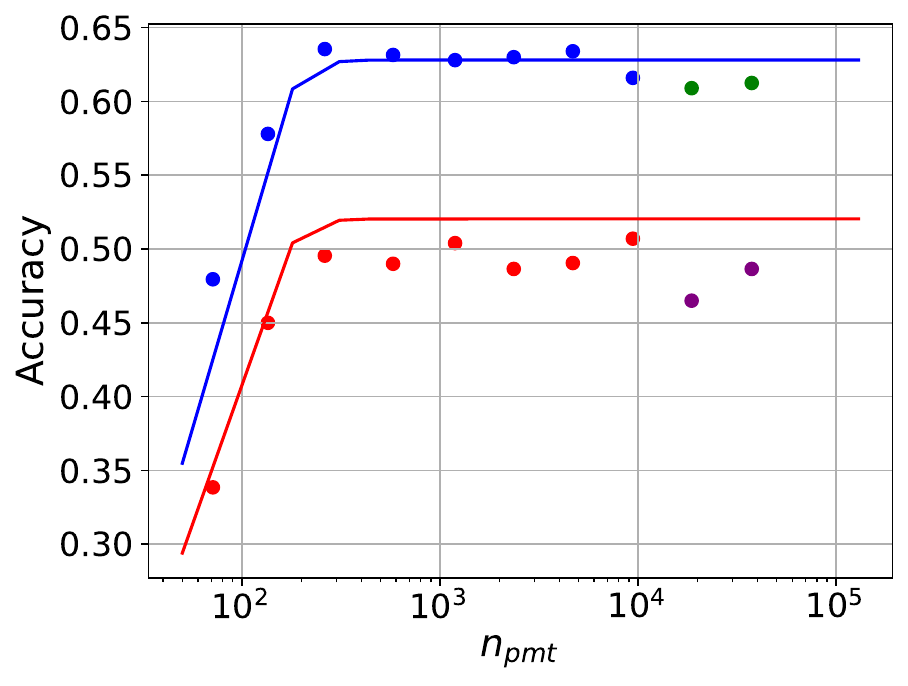}
    \caption{Common sense reasoning.}
    \label{fig:commonsense_reasoning_2d_generalization}
  \end{subfigure}
  \hfill
  \begin{subfigure}[b]{0.31\textwidth}
    \centering
    \includegraphics[width=\textwidth]{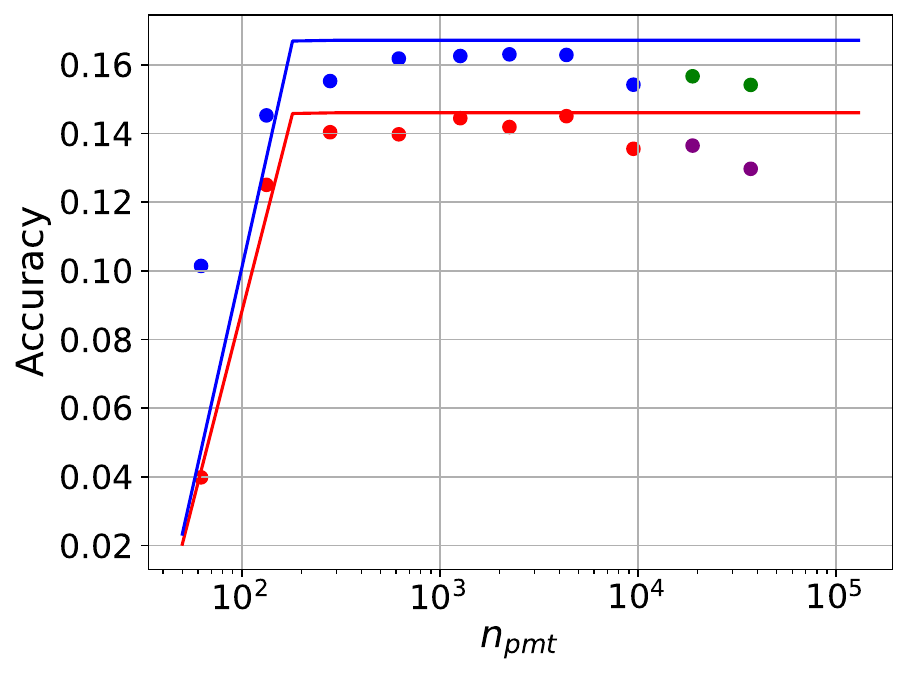}
    \caption{Machine translation.}
    \label{fig:machine_translation_2d_generalization}
  \end{subfigure}
  \caption{Contours of fits at $C=7.8 \times 10^{22}$ (\textcolor{red}{red}) and $C=1.5 \times 10^{23}$ (\textcolor{blue}{blue}) for $n_\text{ctx}=128\text{k}$ on three tasks: arithmetic reasoning (left), common sense reasoning (middle) and machine translation (right). Held-out observations are colored in \textcolor{mlppurple}{purple} and \textcolor{mplgreen}{green} for Llama-2-7b and Llama-2-13b, respectively.}
  \label{fig:downstream-generalization-context}
\end{figure*}

Our results confirm that $\mathcal{P}$ can be jointly determined by the training compute and context length. Increasing $C$ corresponds to an increase in $\mathcal{P}$, in effect shifting up the contour by some diminishing amount in $C$ in the region where $n_\text{pmt} < n_\text{ctx}$. Similarly, increasing $n_\text{pmt}$ when $n_\text{pmt}$ is small leads to significant gains in $\mathcal{P}$, which diminish (sub-linearly for arithmetic reasoning and super-linearly for common sense reasoning and machine translation) and saturate quickly. In the context of our task construction, this makes a lot of sense; the first few in-context task demonstrations go far in improving the generated responses, but once the model has seen enough context to sufficiently capture the task structure, additional demonstrations provide little marginal benefit~\citep{brown2020languagemodelsfewshotlearners}. Additionally, the optimal number of demonstrations is task-dependent; our results suggest that models make better use of additional demonstrations on arithmetic reasoning tasks than they do on common sense reasoning or machine translation tasks.

The remainder of this section aims to study the extent to which our fits generalize to out-of-distribution amounts of training compute (Section~\ref{sec:generalization-compute}), context length (Section~\ref{sec:generalization-context}), and context-extension method (Section~\ref{sec:extension-technique}). Finally, Section~\ref{sec:penalty-ablation} analyzes the role of the sigmoid penalty term.

\subsection{Generalization along \texorpdfstring{$C$}{Lg}}
\label{sec:generalization-compute}

Our scaling laws are fit over a narrow range of $C$, specifically $7.8 \times 10^{22} \leq C \leq 1.5 \times 10^{23}$. To test how well our fits generalize outside of this range, we evaluate several testing models (namely, Qwen2.5-0.5B~\citep{qwen2025qwen25technicalreport}, Gemma-2~\citep{gemmateam2024gemma2improvingopen}, and Llama-2-70B~\citep{touvron2023llama2openfoundation}) ranging between 0.5B to 70B parameters and spanning 3 orders of magnitude in $C$. We evaluate these models at their respective context limits\footnote{While Gemma-2 has a context limit of 8k tokens, it uses sliding window attention for every odd layer. Since this behavior is not supported in vLLM~\citep{kwon2023efficient}, we treat Gemma-2 as if its context limit is 4k tokens.}, and report the prediction error on each task in Table~\ref{tab:results_alt}.

We observe good generalization across these 5 testing models, with many of the prediction errors falling near or below 5 points. Interestingly, our fits generalize the worst to Gemma-2-2B, despite generalizing well to Gemma-2-9B and Gemma-2-27B. Moreover, we achieve stronger generalization on arithmetic reasoning and machine translation tasks compared to common sense reasoning, which aligns with our in-distribution results. Finally, these results suggest we tend to underestimate the performance when $C$ is small, and slightly overestimate performance when $C$ is large.

\subsection{Generalization along \texorpdfstring{$n_\text{pmt}$}{Lg}}
\label{sec:generalization-context}

In order to measure how well our scaling laws generalize to longer contexts, we refit our scaling curves, this time holding out observations where the context length exceeds 10,000 tokens. Figure~\ref{fig:downstream-generalization-context} displays contours of our fits at $C=7.8 \times 10^{22}$ and $C=1.5 \times 10^{23}$ for $n_\text{ctx} = 128\text{k}$ for each task. Again, we see strong generalization along $n_\text{pmt}$, achieving prediction errors of just 0.017, 0.067, and 0.006 across the held-out observations on arithmetic reasoning, common sense reasoning, and machine translation, respectively. These low error rates across diverse tasks demonstrate that our joint scaling framework can reliably extrapolate to longer context lengths, making it particularly suitable for long-context LLM design.

Interestingly, on common sense reasoning and machine translation tasks, we observe that $\mathcal{P}$ is inversely proportional to $n_\text{ctx}$ for some fixed $n_\text{pmt}$. That is, as we extend the context, performance slightly worsens. We hypothesize that this decline is not due to an intrinsic scaling trend but rather because the training mix used to extend the context is misaligned with these tasks. For example, our training mix is sourced from PG-19~\citep{rae2019compressivetransformerslongrangesequence}, which includes predominantly English text, so it's unsurprising that machine translation performance worsens with increased training. 

\begin{table*}[htbp]
    \small
    \centering
    \begin{tabular}{lccccc}
    \toprule
    Model & $C$ & $n_\text{pmt}$ & $\mathcal{P}_{\text{AR}} - \hat{\mathcal{P}}_{\text{AR}}$ & $\mathcal{P}_{\text{CSR}} - \hat{\mathcal{P}}_{\text{CSR}}$ & $\mathcal{P}_{\text{MT}} - \hat{\mathcal{P}}_{\text{MT}}$\\
    \midrule
    Llama-2-7B (PI)   & $7.777 \times 10^{22}$ & 32k & +0.014 & +0.079 & -0.005 \\
    Llama-2-7B (YaRN) & $7.775 \times 10^{22}$ & 32k & +0.005 & +0.014 & -0.005 \\
    \bottomrule
    \end{tabular}
    \caption{Generalization of fit on test models for arithmetic reasoning (AR), common sense reasoning (CSR), and machine translation (MT) at $n_\text{ctx}=32\text{k}$. }
    \label{tab:pi_vs_yarn}
\end{table*}

\begin{table*}[htbp]
  \small
  \centering
  \begin{tabular}{lccc}
    \toprule
     & $|P-\hat{P}|_{n_\text{pmt} \leq n_\text{ctx}}$ & $|P-\hat{P}|_{n_\text{pmt} > n_\text{ctx}}$ & $|P-\hat{P}|$ \\
    \midrule
    With penalty term    & 0.010 & 0.014 & 0.010 \\
    Without penalty term & 0.019 & 0.104 & 0.029 \\
    \bottomrule
  \end{tabular}
  \caption{Prediction errors on the arithmetic reasoning task, with and without the sigmoid penalty term.}
  \label{tab:penalty-ablation}
\end{table*}

\subsection{Does the choice of context extension technique matter?} 
\label{sec:extension-technique}

A number of different techniques have been proposed for extending the context length of a model with rotary positional embeddings~\citep{chen2023extendingcontextwindowlarge, peng2023yarnefficientcontextwindow, xiong2023effectivelongcontextscalingfoundation}. It's natural to wonder how sensitive our fit scaling curves are to one's choice of context extension technique. To test this, we evaluate Together's Llama-2-7B model~\citep{together} extended to 32k context via positional interpolation~\citep{chen2023extendingcontextwindowlarge}. We evaluate this model at its context limit of 32k tokens across our 3 tasks and compute the prediction error for each, that is, the difference between the observed performance $\mathcal{P}$ and the predicted performance $\hat{\mathcal{P}}$. We compare against the prediction error on the Llama-2-7B checkpoint extended to $n_\text{ctx}=32\text{k}$ via YaRN~\citep{peng2023yarnefficientcontextwindow}. It's worth noting that the training mix and quantity are different between these two models; Together's was trained on 1.5B tokens of a diverse data mix, while we follow~\citet{peng2023yarnefficientcontextwindow} and train on just 0.836B tokens from PG-19~\citep{rae2019compressivetransformerslongrangesequence}. Still, our compute estimates for both models are sufficiently similar ($7.777 \times 10^{22}$ vs $7.775 \times 10^{22}$ FLOPs, respectively). 

Table~\ref{tab:pi_vs_yarn} lists the results. In general, the prediction errors we observe on Together's Llama-2-7B model extended via positional interpolation are similar to the prediction errors we observe on our Llama-2-7B model extended via YaRN. These results suggest that the choice of context extension technique has little impact on the scaling properties of downstream performance.

\subsection{Ablation over the penalty term}
\label{sec:penalty-ablation}
To quantify the impact of our sigmoid penalty for prompt lengths exceeding the model’s context limit, we fit Eq.~\eqref{eq:1} on the arithmetic reasoning task with and without the penalty term. Table~\ref{tab:penalty-ablation} reports the resulting prediction errors. We observe that without the penalty term, the fit underestimates performance when $n_\text{pmt} \leq n_\text{ctx}$ and overestimates performance when $n_\text{pmt} > n_\text{ctx}$, confirming the importance of the penalty term.

\section{Related Work}
\citet{hestness2017deeplearningscalingpredictable} and~\citet{kaplan2020scalinglawsneurallanguage} introduce scaling laws which describe the relationship between upstream model performance (e.g., cross-entropy loss) and model design features (e.g., the number of model parameters, the size of the training dataset, or the total amount of training compute).~\citet{henighan2020scalinglawsautoregressivegenerative} extends this analysis to other types of autoregressive models (e.g., generative image and video modeling).~\citet{hoffmann2022trainingcomputeoptimallargelanguage} and~\citet{openai2024gpt4technicalreport} describe the use of scaling laws to train compute-optimal LLMs, and~\citet{caballero2023brokenneuralscalinglaws} introduces a form of smoothly broken neural scaling laws to better capture non-monotonic scaling.

Several works have focused on scaling laws for predicting downstream performance.~\citet{wei2022emergentabilitieslargelanguage} and~\citet{hu2024predictingemergentabilitiesinfinite} focus on predicting abilities that ``emerge'' in LLMs when trained on enough compute.~\citet{isik2024scalinglawsdownstreamtask} explores scaling laws for transfer learning on machine translation tasks, while~\citet{schaeffer2025predictingdownstreamcapabilitiesfrontier} studies scaling laws for downstream multiple-choice tasks. Other works have employed a collaborative approach and source performance data from public benchmarks to better generalize across different model families~\citep{zhang2024collaborativeperformancepredictionlarge, ruan2024observationalscalinglawspredictability, polo2025slothscalinglawsllm, gadre2024languagemodelsscalereliably}.~\citet{chen2024scalinglawspredictingdownstream} and~\citet{ruan2024observationalscalinglawspredictability} employ a two-stage approach, using an intermediary (e.g., upstream loss) for predicting downstream performance. Both~\citet{owen2024predictablelanguagemodelbenchmark} and~\citet{ye2023predictablelargelanguagemodel} aim to predict aggregate performance on benchmarks such as BIG-Bench~\citep{srivastava2023imitationgamequantifyingextrapolating}. Comparatively, this work introduces a dependence on the context length and suggests that you can predict downstream performance and obtain strong generalization (even across model families) with a straightforward, interpretable functional form. 

Context refers to the information provided to a model at inference time, such as few-shot demonstrations~\citep{brown2020languagemodelsfewshotlearners}, retrieved evidence~\citep{lewis2020retrieval}, or task instructions~\citep{crispino2024agent}. Though context shapes performance by conditioning on structured or unstructured information, little scaling analysis has been conducted on the role of context. Both~\citet{kaplan2020scalinglawsneurallanguage} and~\citet{caballero2023brokenneuralscalinglaws} briefly explore the scaling of upstream performance as it relates to context length.~\citet{xiong2023effectivelongcontextscalingfoundation} extends the context limit of Llama-2 and finds that validation loss scales as a power law in the context length, but stops short of exploring the relationship between downstream performance and context length.~\citet{caballero2023brokenneuralscalinglaws} and~\citet{brown2020languagemodelsfewshotlearners} explore the diminishing returns of increasing the number of in-context demonstrations. To the best of our knowledge, our work is the first to explicitly focus on the scaling relationship between downstream performance and context length, and the first attempt to unify the understanding of scaling with respect to both context and compute.    

The ability of an LLM to extrapolate to longer sequences depends heavily on its positional encodings. While some positional encoding techniques (e.g., ALiBi~\citep{press2022trainshorttestlong}) offer limited length extrapolation, other common techniques (e.g., RoPE~\citep{su2023roformerenhancedtransformerrotary}) don't. As a result, a number of techniques to efficiently extend the context window of LLMs have been proposed.

Some techniques offer training-free context extension, typically by adjusting the attention mechanism itself.~\citet{jin2024llmmaybelonglmselfextend} leverages a bi-level attention mechanism, applying standard self-attention to adjacent tokens and grouped attention for distant tokens. InfLLM is a memory‐based technique that integrates sliding-window attention with block-level context memory~\citep{xiao2024infllmtrainingfreelongcontextextrapolation}. Similarly, LM-Infinite employs a $\Lambda$-shaped attention mask, effectively masking attention over tokens in the middle, and restricts the maximum positional difference between any two tokens to the maximum sequence length seen during pre-training~\citep{han2024lminfinitezeroshotextremelength}. On the other hand,~\citet{10.5555/3692070.3692130} introduces dual-chunk attention, which decomposes the attention computation into chunk-based modules to better capture the relative positional information between distant tokens.

Additionally, a number of techniques have been proposed that focus on rescaling the positional encodings. Concurrently,~\citet{chen2023extendingcontextwindowlarge} and~\citet{kaiokendev} introduced position interpolation, which extends the context window by linearly interpolating the position indices to be within the pre-trained context limit.~\citet{xiong2023effectivelongcontextscalingfoundation} proposes decreasing the rotational angle (base frequency) of RoPE to prevent the relative positional information from decaying. Building on this, NTK-aware interpolation~\citep{blocntkaware} adjusts the scaling for each RoPE dimension based on its frequency, thereby mitigating the loss of high-frequency details.~\citet{blocntkparts} introduces NTK-by-parts interpolation, which selectively interpolates lower-frequency dimensions while preserving higher-frequency components to maintain local relative positioning. YaRN~\citep{peng2023yarnefficientcontextwindow} combines NTK-by-parts with a mechanism to rescale the logits in the attention softmax to further improve performance on long sequences. In this work, we utilize YaRN to extend the context limit of the Llama-2 models due to its high compute efficiency and strong empirical results compared to other techniques.

\section{Conclusion}
In this work, we introduce a straightforward, interpretable framework that jointly models downstream performance as a function of the training compute and the provided context. Extensive experiments on arithmetic reasoning, common-sense reasoning, and machine translation tasks demonstrate that our framework not only fits the in-distribution performance accurately but also generalizes well across 3 orders of magnitude in the amount of non-embedding training compute $C$, 4 orders of magnitude in the amount of input context length, and even to other context-extension techniques. These findings reveal that downstream performance benefits from increased compute and longer, relevant context, but only up to a saturation point. Our work thus provides actionable insights for designing more effective long-context LLMs and bridges the gap between upstream scaling metrics and real-world task performance.

\section*{Acknowledgments}
This work was supported in part by a research gift from Google.

\section*{Limitations}
While our proposed context-aware scaling framework provides an interpretable approach to modeling downstream performance, it does come with limitations. Specifically, our formulation relies on a set of assumptions (e.g., performance scales with training compute and context) that may not hold under extreme scaling regimes or in the presence of adversarial attacks like many-shot jailbreaking~\citep{NEURIPS2024_ea456e23}. Moreover, factors such as the pre-training data mix, post-training and alignment, and architectural choices, which can all influence downstream model performance, are not explicitly accounted for. However, these factors likely affect the optimal parameters of a fit without necessarily changing the structure of Eq.~\eqref{eq:1}. For example, post-training alignment (e.g., instruction tuning) might improve a model's zero-shot performance, resulting in a higher value for the parameter $A$ compared to a non-aligned base model. Future work could investigate how these factors and others influence the identified parameters, enhancing the framework's predictive power while retaining its interpretable form. Lastly, our scaling curves are fit to a narrow range of training compute, and may fail to generalize well to LLMs trained on an amount of compute that extends far beyond this range.

\bibliography{custom}

\appendix
\clearpage
\section{Dataset Details} 
\label{sec:dataset_details}
\paragraph{GSM8K~\citep{cobbe2021trainingverifierssolvemath}} We filter out instances over 256 tokens in length, and select 511 training instances and 250 testing instances at random. During inference, we allow up to 400 new tokens. The average token lengths of the training and testing instances were 177.64 and 177.43 respectively. The generated responses averaged around 172.13 tokens in length. To evaluate, we extract the model's final answer and compare it with the reference answer, checking for numerical equivalence.
\paragraph{MATH~\citep{hendrycks2021measuringmathematicalproblemsolving}}  We filter out instances over 256 tokens in length, and select 511 training instances and 250 testing instances at random. During inference, we allow up to 400 new tokens. The average token lengths of the training and testing instances were 160.54 and 155.74 respectively. The generated responses also averaged around 184.0 tokens in length. To evaluate, we extract the model's final answer and compare it with the reference answer, checking for numerical equivalence.
\paragraph{AQUA-RAT~\citep{ling2017programinductionrationalegeneration}} We filter out instances over 256 tokens in length, and select 511 training instances and 250 testing instances at random. We allow up to 5 new tokens during generation. The average token lengths of the training and testing instances were 88.45 and 93.09 respectively. The generated responses also averaged around 3.44 tokens in length. To evaluate, we check to see if the choice returned by our model matches the reference answer.
\paragraph{DeepMind Math~\citep{saxton2019analysingmathematicalreasoningabilities}} The dataset is categorized into 56 subsets. We filter out instances over 256 tokens in length, and select 511 training instances and 50 testing instances at random from each subset. We allow up to 400 new tokens during generation. The average token lengths of the training and testing instances were 57.94 and 61.05 respectively. The generated responses also averaged around 85.71 tokens in length. To evaluate, we extract the model's final answer and compare it with the reference answer, checking for numerical equivalence.
\paragraph{PIQA~\citep{bisk2019piqareasoningphysicalcommonsense}} We filter out instances over 256 tokens in length, and select 511 training instances and 250 testing instances at random. We allow up to 5 new tokens during generation. The average token lengths of the training and testing instances were 81.16 and 81.55 respectively. The generated responses also averaged around 3.46 tokens in length. To evaluate, we check to see if the choice returned by our model matches the reference answer.
\paragraph{OpenBookQA~\citep{mihaylov2018suitarmorconductelectricity}} We filter out instances over 256 tokens in length, and select 511 training instances and 250 testing instances at random. We allow up to 5 new tokens during generation. The average token lengths of the training and testing instances were 47.74 and 49.39 respectively. The generated responses also averaged around 3.3 tokens in length. To evaluate, we check to see if the choice returned by our model matches the reference answer.
\paragraph{SIQA~\citep{sap2019socialiqacommonsensereasoningsocial}} We filter out instances over 256 tokens in length, and select 511 training instances and 250 testing instances at random. We allow up to 1 new token during generation. The average token lengths of the training and testing instances were 56.68 and 56.87 respectively. The generated responses also averaged around 3.35 tokens in length. To evaluate, we check to see if the choice returned by our model matches the reference answer.
\paragraph{HellaSwag~\citep{zellers2019hellaswagmachinereallyfinish}} We filter out instances over 256 tokens in length, and select 511 training instances and 250 testing instances at random. We allow up to 5 new tokens during generation. The average token lengths of the training and testing instances were 153.06 and 156.05 respectively. The generated responses also averaged around 3.67 tokens in length. To evaluate, we check to see if the choice returned by our model matches the reference answer.
\paragraph{WinoGrande~\citep{sakaguchi2019winograndeadversarialwinogradschema}} We filter out instances over 256 tokens in length, and select 511 training instances and 250 testing instances at random. We allow up to 5 new tokens during generation. The average token lengths of the training and testing instances were 53.98 and 53.87 respectively. The generated responses also averaged around 3.33 tokens in length. To evaluate, we check to see if the choice returned by our model matches the reference answer.
\paragraph{ARC Easy~\citep{clark2018thinksolvedquestionanswering}} We filter out instances over 256 tokens in length, and select 511 training instances and 250 testing instances at random. We allow up to 5 new tokens during generation. The average token lengths of the training and testing instances were 66.69 and 67.14 respectively. The generated responses also averaged around 3.46 tokens in length. To evaluate, we check to see if the choice returned by our model matches the reference answer.
\paragraph{ARC Challenge~\citep{clark2018thinksolvedquestionanswering}} We filter out instances over 256 tokens in length, and select 511 training instances and 250 testing instances at random. We allow up to 5 new tokens during generation. The average token lengths of the training and testing instances were 75.65 and 76.83 respectively. The generated responses also averaged around 3.43 tokens in length. To evaluate, we check to see if the choice returned by our model matches the reference answer.
\paragraph{CommenSenseQA~\citep{talmor2019commonsenseqaquestionansweringchallenge}} We filter out instances over 256 tokens in length, and 250 select 511 training instances and testing instances at random. We allow up to 5 new tokens during generation. The average token lengths of the training and testing instances were 50.42 and 49.92 respectively. The generated responses also averaged around 1.0 tokens in length. To evaluate, we check to see if the choice returned by our model matches the reference answer.
\paragraph{WMT14 (CS-EN)~\citep{bojar-EtAl:2014:W14-33}} We filter out instances over 256 tokens in length, and select 511 training instances and 250 testing instances at random. We allow up to 256 new tokens during generation. The average token lengths of the training and testing instances were 95.01 and 85.25 respectively. The generated responses also averaged around 77.77 tokens in length. We use BLEU-4~\citep{10.3115/1073083.1073135} to score the generated translations relative to the reference translations.
\paragraph{WMT14 (DE-EN)~\citep{bojar-EtAl:2014:W14-33}} We filter out instances over 256 tokens in length, and select 511 training instances and 250 testing instances at random. We allow up to 256 new tokens during generation. The average token lengths of the training and testing instances were 85.53 and 77.68 respectively. The generated responses also averaged around 77.77 tokens in length. We use BLEU-4~\citep{10.3115/1073083.1073135} to score the generated translations relative to the reference translations.
\paragraph{WMT14 (FR-EN)~\citep{bojar-EtAl:2014:W14-33}} We filter out instances over 256 tokens in length, and select 511 training instances and 250 testing instances at random. We allow up to 256 new tokens during generation. The average token lengths of the training and testing instances were 95.94 and 84.29 respectively. The generated responses also averaged around 78.73 tokens in length. We use BLEU-4~\citep{10.3115/1073083.1073135} to score the generated translations relative to the reference translations.
\paragraph{WMT14 (HI-EN)~\citep{bojar-EtAl:2014:W14-33}} We filter out instances over 256 tokens in length, and select 511 training instances and 250 testing instances at random. We allow up to 256 new tokens during generation. The average token lengths of the training and testing instances were 34.01 and 147.09 respectively. The generated responses also averaged around 53.11 tokens in length. We use BLEU-4~\citep{10.3115/1073083.1073135} to score the generated translations relative to the reference translations.
\paragraph{WMT14 (RU-EN)~\citep{bojar-EtAl:2014:W14-33}} We filter out instances over 256 tokens in length, and select 511 training instances and 250 testing instances at random. We allow up to 256 new tokens during generation. The average token lengths of the training and testing instances were 73.54 and 86.56 respectively. The generated responses also averaged around 77.24 tokens in length. We use BLEU-4~\citep{10.3115/1073083.1073135} to score the generated translations relative to the reference translations.

\section{Full Results}
\label{sec:full-results}
\begin{figure*}
  \centering
  \begin{subfigure}[b]{0.45\textwidth}
    \centering
    \includegraphics[width=\textwidth]{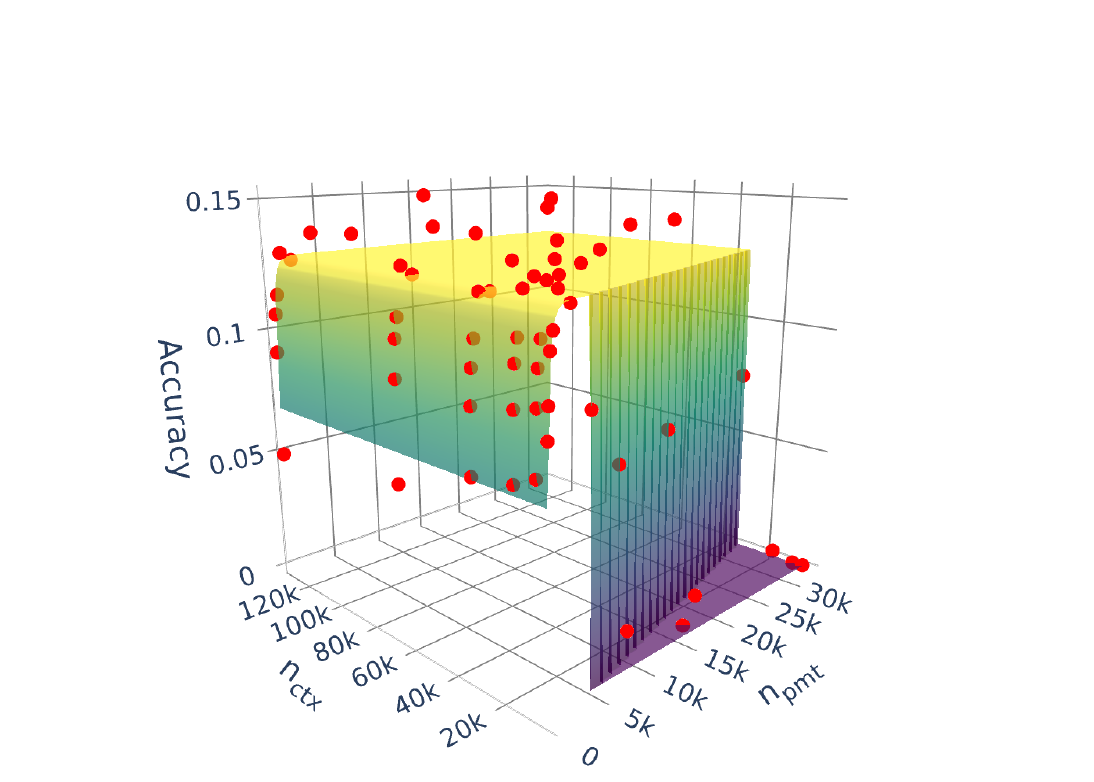}
  \end{subfigure}
  \hfill
  \begin{subfigure}[b]{0.45\textwidth}
    \centering
    \includegraphics[width=\textwidth]{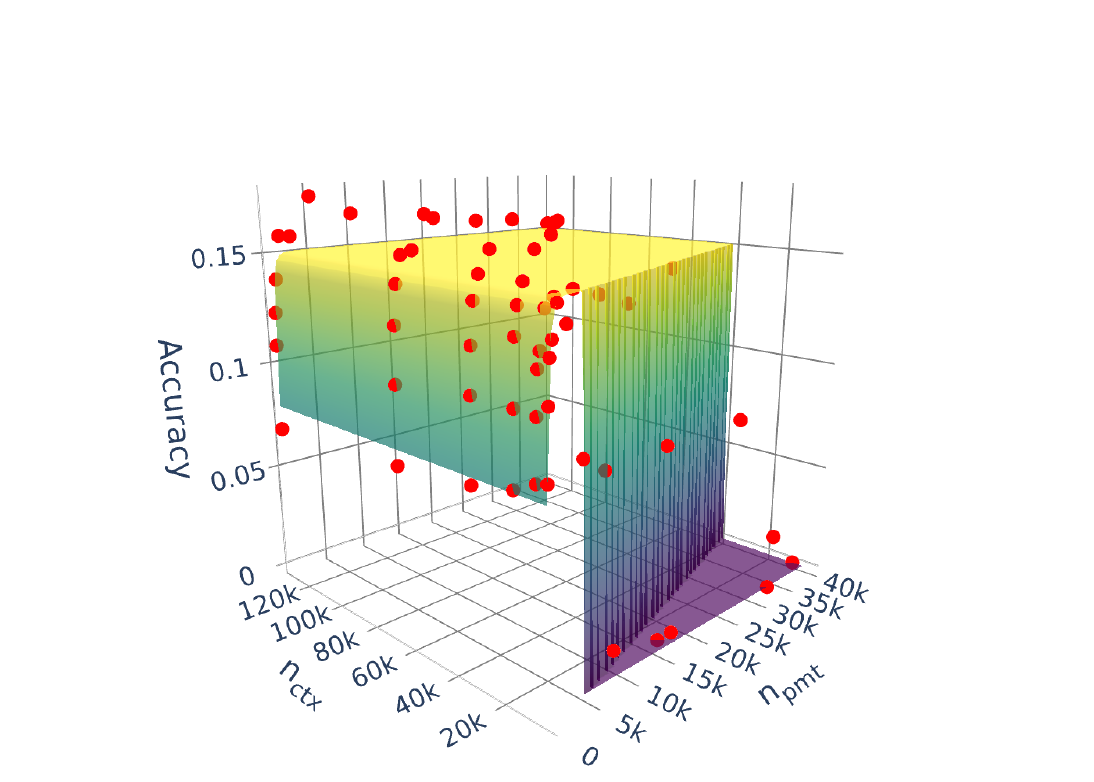}
  \end{subfigure}
  \caption{Contours of our fit at $C=7.8 \times 10^{22}$ (left) and $C=1.5 \times 10^{23}$ (right) for the arithmetic reasoning task.}
  \label{fig:contours-arithmetic}
\end{figure*}

\begin{figure*}
  \centering
  \begin{subfigure}[b]{0.45\textwidth}
    \centering
    \includegraphics[width=\textwidth]{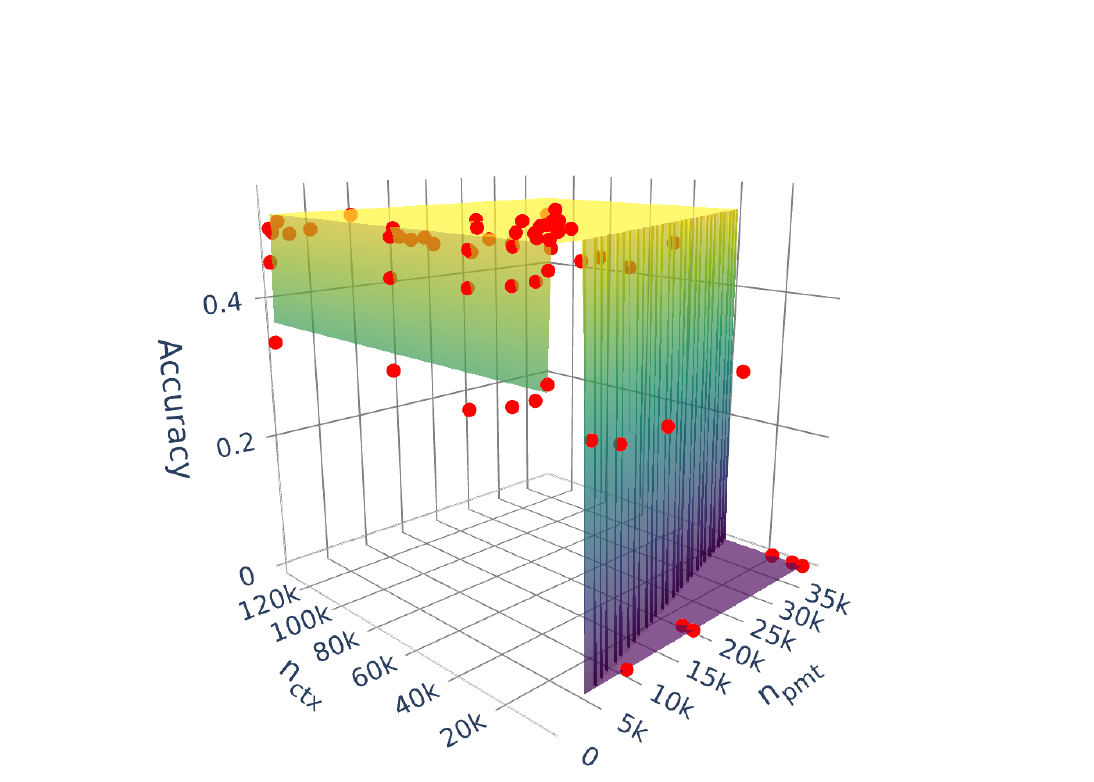}
  \end{subfigure}
  \hfill
  \begin{subfigure}[b]{0.45\textwidth}
    \centering
    \includegraphics[width=\textwidth]{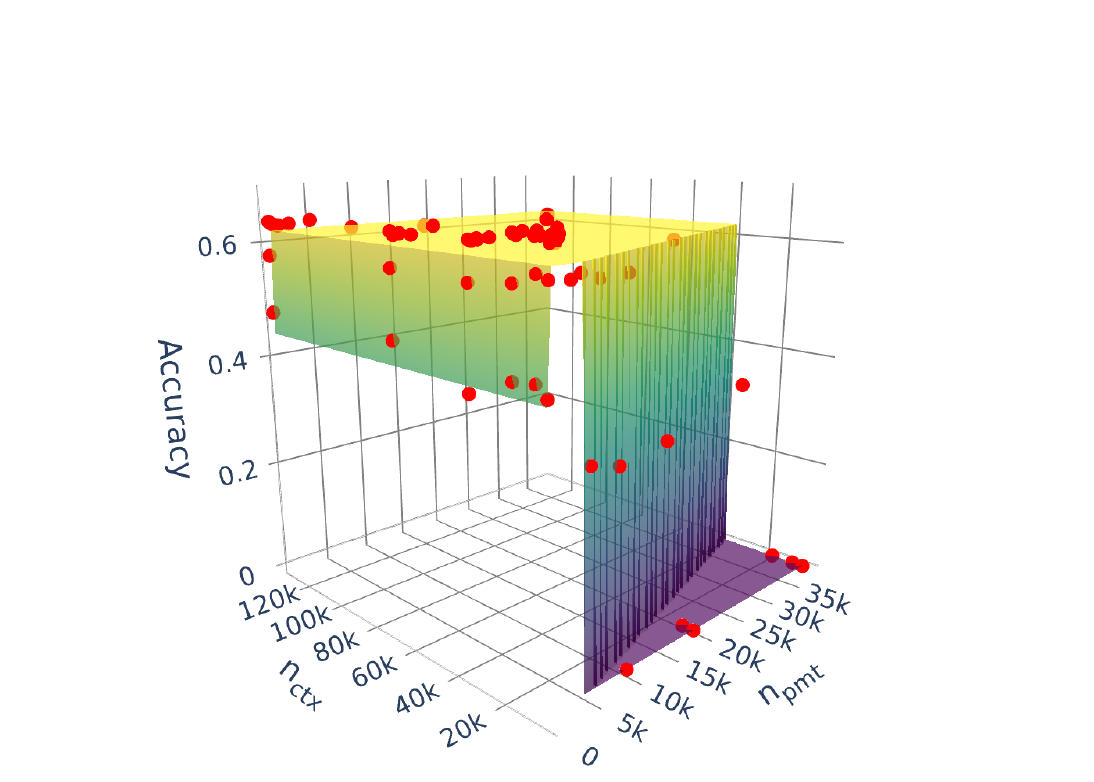}
  \end{subfigure}
  \caption{Contours of our fit at $C=7.8 \times 10^{22}$ (left) and $C=1.5 \times 10^{23}$ (right) for the common sense reasoning task.}
  \label{fig:contours-commonsense}
\end{figure*}

\begin{figure*}
  \centering
  \begin{subfigure}[b]{0.45\textwidth}
    \centering
    \includegraphics[width=\textwidth]{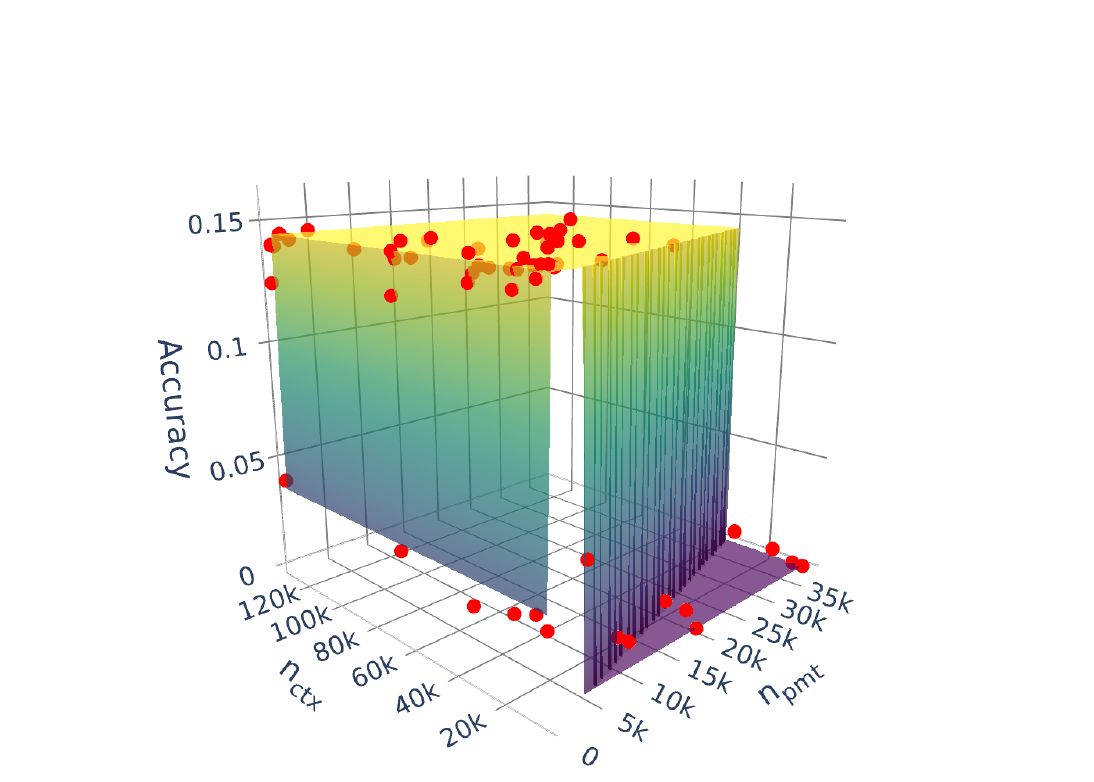}
  \end{subfigure}
  \hfill
  \begin{subfigure}[b]{0.45\textwidth}
    \centering
    \includegraphics[width=\textwidth]{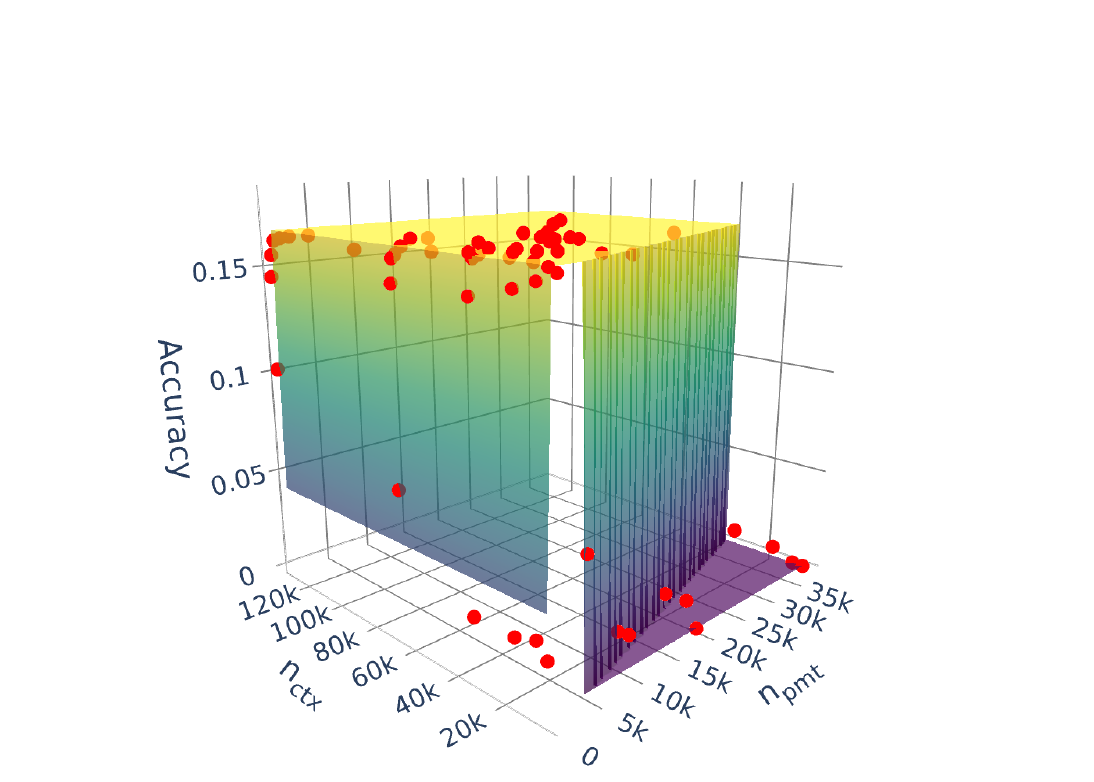}
  \end{subfigure}
  \caption{Contours of our fit at $C=7.8 \times 10^{22}$ (left) and $C=1.5 \times 10^{23}$ (right) for the machine translation task.}
  \label{fig:contours-translation}
\end{figure*}

In this section, we present full aggregate results in Tables~\ref{tab:arithmeticreasoning},~\ref{tab:commonsensereasoning}, and~\ref{tab:machinetranslation} for arithmetic reasoning, common sense reasoning, and machine translation respectively. Figures~\ref{fig:contours-arithmetic},~\ref{fig:contours-commonsense},and~\ref{fig:contours-translation} provide contours of our fits at $C=7.8 \times 10^{22}$ and $C=1.5 \times 10^{23}$.

\begin{table*}
    \centering
    \scriptsize
    \begin{tabular}{lcccccccccc}
    \toprule
    $k$ & 0 shots & 1 shot & 3 shots & 7 shots & 15 shots & 31 shots & 63 shots & 127 shots & 255 shots & 511 shots \\ \midrule
    {Llama-2-7b-hf} & {0.089} & {0.099} & {0.115} & {0.120} & {0.136} & {0.127} & {0.094} & {0.014} & {0.014} & {0.000} \\
    {Yarn-Llama-2-7b-8k} & {0.076} & {0.097} & {0.109} & {0.117} & {0.134} & {0.131} & {0.137} & {0.071} & {0.000} & {0.000} \\
    {Yarn-Llama-2-7b-16k} & {0.072} & {0.095} & {0.109} & {0.116} & {0.130} & {0.133} & {0.143} & {0.139} & {0.073} & {0.002} \\
    {Yarn-Llama-2-7b-32k} & {0.069} & {0.092} & {0.104} & {0.113} & {0.127} & {0.127} & {0.135} & {0.134} & {0.143} & {0.076} \\
    {Yarn-Llama-2-7b-64k} & {0.057} & {0.094} & {0.108} & {0.115} & {0.132} & {0.128} & {0.143} & {0.140} & {0.150} & {0.138} \\
    {Yarn-Llama-2-7b-128k} & {0.049} & {0.091} & {0.106} & {0.113} & {0.129} & {0.126} & {0.136} & {0.135} & {0.149} & {0.140} \\ \midrule
    {Llama-2-13b-hf} & {0.088} & {0.115} & {0.131} & {0.137} & {0.148} & {0.141} & {0.092} & {0.011} & {0.005} & {0.000} \\
    {Yarn-Llama-2-13b-8k} & {0.086} & {0.110} & {0.126} & {0.132} & {0.146} & {0.149} & {0.151} & {0.082} & {0.000} & {0.000} \\
    {Yarn-Llama-2-13b-16k} & {0.081} & {0.110} & {0.135} & {0.146} & {0.153} & {0.163} & {0.172} & {0.145} & {0.077} & {0.010} \\
    {Yarn-Llama-2-13b-32k} & {0.077} & {0.111} & {0.129} & {0.145} & {0.154} & {0.162} & {0.171} & {0.169} & {0.134} & {0.065} \\
    {Yarn-Llama-2-13b-64k} & {0.073} & {0.106} & {0.130} & {0.146} & {0.156} & {0.158} & {0.169} & {0.167} & {0.159} & {0.136} \\
    {Yarn-Llama-2-13b-128k} & {0.069} & {0.108} & {0.123} & {0.138} & {0.157} & {0.157} & {0.174} & {0.165} & {0.163} & {0.153} \\ \bottomrule
    \end{tabular}
    \caption{Accuracy on arithmetic reasoning, aggregated over every instance in the task.}
    \label{tab:arithmeticreasoning}
\end{table*}

\begin{table*}
    \centering
    \scriptsize
    \begin{tabular}{lcccccccccc}
    \toprule
    $k$ & 0 shots & 1 shot & 3 shots & 7 shots & 15 shots & 31 shots & 63 shots & 127 shots & 255 shots & 511 shots \\ \midrule
    {Llama-2-7b-hf}          & {0.376} & {0.489} & {0.518} & {0.536} & {0.536} & {0.527} & {0.302} & {0.000} & {0.000} & {0.000} \\
    {Yarn-Llama-2-7b-8k}     & {0.356} & {0.476} & {0.518} & {0.530} & {0.530} & {0.523} & {0.491} & {0.278} & {0.000} & {0.000} \\
    {Yarn-Llama-2-7b-16k}    & {0.342} & {0.468} & {0.508} & {0.522} & {0.532} & {0.519} & {0.521} & {0.486} & {0.264} & {0.000} \\
    {Yarn-Llama-2-7b-32k}    & {0.325} & {0.459} & {0.500} & {0.496} & {0.522} & {0.508} & {0.501} & {0.534} & {0.457} & {0.276} \\
    {Yarn-Llama-2-7b-64k}    & {0.346} & {0.456} & {0.503} & {0.513} & {0.502} & {0.498} & {0.490} & {0.515} & {0.470} & {0.458} \\
    {Yarn-Llama-2-7b-128k}   & {0.338} & {0.450} & {0.496} & {0.490} & {0.504} & {0.486} & {0.490} & {0.507} & {0.465} & {0.486} \\ \midrule
    {Llama-2-13b-hf}         & {0.453} & {0.604} & {0.649} & {0.660} & {0.659} & {0.600} & {0.344} & {0.000} & {0.000} & {0.000} \\
    {Yarn-Llama-2-13b-8k}    & {0.469} & {0.610} & {0.662} & {0.656} & {0.675} & {0.652} & {0.603} & {0.318} & {0.000} & {0.000} \\
    {Yarn-Llama-2-13b-16k}   & {0.464} & {0.594} & {0.656} & {0.654} & {0.658} & {0.650} & {0.658} & {0.584} & {0.308} & {0.000} \\
    {Yarn-Llama-2-13b-32k}   & {0.432} & {0.586} & {0.642} & {0.640} & {0.642} & {0.642} & {0.646} & {0.626} & {0.567} & {0.322} \\
    {Yarn-Llama-2-13b-64k}   & {0.481} & {0.589} & {0.642} & {0.636} & {0.638} & {0.634} & {0.645} & {0.620} & {0.614} & {0.582} \\
    {Yarn-Llama-2-13b-128k}  & {0.480} & {0.578} & {0.636} & {0.632} & {0.628} & {0.630} & {0.634} & {0.616} & {0.609} & {0.612} \\ \bottomrule
    \end{tabular}
    \caption{Accuracy on Commonsense Reasoning tasks, aggregated over every instance in the task.}
    \label{tab:commonsensereasoning}
\end{table*}

\begin{table*}
    \centering
    \scriptsize
    \begin{tabular}{lcccccccccc}
    \toprule
    $k$ & 0 shots & 1 shot & 3 shots & 7 shots & 15 shots & 31 shots & 63 shots & 127 shots & 255 shots & 511 shots \\ \midrule
    {Llama-2-7b-hf}          & {0.031} & {0.147} & {0.155} & {0.154} & {0.156} & {0.159} & {0.049} & {0.011} & {0.000} & {0.000} \\
    {Yarn-Llama-2-7b-8k}     & {0.034} & {0.142} & {0.155} & {0.146} & {0.151} & {0.152} & {0.152} & {0.010} & {0.006} & {0.000} \\
    {Yarn-Llama-2-7b-16k}    & {0.031} & {0.138} & {0.152} & {0.144} & {0.147} & {0.144} & {0.143} & {0.143} & {0.006} & {0.003} \\
    {Yarn-Llama-2-7b-32k}    & {0.026} & {0.138} & {0.147} & {0.141} & {0.143} & {0.142} & {0.141} & {0.140} & {0.146} & {0.005} \\
    {Yarn-Llama-2-7b-64k}    & {0.033} & {0.129} & {0.144} & {0.142} & {0.148} & {0.141} & {0.148} & {0.142} & {0.144} & {0.134} \\
    {Yarn-Llama-2-7b-128k}   & {0.040} & {0.125} & {0.140} & {0.140} & {0.144} & {0.142} & {0.145} & {0.136} & {0.136} & {0.130} \\ \midrule
    {Llama-2-13b-hf}         & {0.023} & {0.166} & {0.175} & {0.180} & {0.181} & {0.175} & {0.058} & {0.015} & {0.000} & {0.000} \\
    {Yarn-Llama-2-13b-8k}    & {0.029} & {0.161} & {0.171} & {0.175} & {0.177} & {0.170} & {0.174} & {0.014} & {0.011} & {0.000} \\
    {Yarn-Llama-2-13b-16k}   & {0.025} & {0.157} & {0.170} & {0.171} & {0.176} & {0.166} & {0.173} & {0.166} & {0.010} & {0.005} \\
    {Yarn-Llama-2-13b-32k}   & {0.025} & {0.152} & {0.168} & {0.166} & {0.171} & {0.168} & {0.164} & {0.156} & {0.160} & {0.007} \\
    {Yarn-Llama-2-13b-64k}   & {0.066} & {0.152} & {0.162} & {0.163} & {0.166} & {0.169} & {0.163} & {0.160} & {0.162} & {0.160} \\
    {Yarn-Llama-2-13b-128k}  & {0.101} & {0.145} & {0.155} & {0.162} & {0.163} & {0.163} & {0.163} & {0.154} & {0.157} & {0.154} \\ \bottomrule
    \end{tabular}
    \caption{Accuracy on Machine Translation tasks, aggregated over every instance in the task.}
    \label{tab:machinetranslation}
\end{table*}

\clearpage

\end{document}